\documentclass{article}
\usepackage{arxiv}
\usepackage{moreverb,url}
\usepackage[utf8]{inputenc} 
\usepackage[T1]{fontenc}    
\usepackage{url}            
\usepackage{booktabs}       
\usepackage{amsfonts}       
\usepackage{nicefrac}       
\usepackage{microtype}      

\usepackage{amssymb}
\usepackage{mathtools}
\usepackage{color}
\usepackage{float}

	\usepackage[pdftex]{graphicx}
	\graphicspath{{Figures/}}
	\usepackage[caption=false,font=footnotesize]{subfig}
	\usepackage{dblfloatfix} 
	\usepackage{array}
	\usepackage{multirow}
	\usepackage{booktabs}
	\usepackage{adjustbox}
	\usepackage{algorithm}
	\usepackage{algorithmicx}
	\usepackage[noend]{algpseudocode}
		\algrenewcommand\algorithmicindent{2.0em}%
		\algnewcommand\Let[2]{\State #1 $\gets$ #2}
		\algnewcommand\AND{\ \textbf{and}\ }
		\algnewcommand\OR{\ \textbf{or} \ }
		\algnewcommand\algorithmicinput{\textbf{Input:}}
		\algnewcommand\Input{\item[\algorithmicinput]}
		\algnewcommand\algorithmiccompute{\textbf{Compute:}}
		\algnewcommand\Compute{\item[\algorithmiccompute]}
		\algnewcommand\algorithmicoutput{\textbf{Output:}}
		\algnewcommand\Output{\item[\algorithmicoutput]}
		\algblockdefx{MRepeat}{EndRepeat}{\textbf{repeat}}{}
		\algnotext{EndRepeat}

\usepackage[backend = biber, style=ieee, url=false]{biblatex}
\DeclareFieldFormat
  [article]
  {journaltitle}{\textit{#1}}
\let\cite\parencite
\usepackage[colorlinks,bookmarksopen,bookmarksnumbered,citecolor=red,urlcolor=red]{hyperref}

	\newcommand{\II}{\ensuremath{\mathbb I}}
	
	\newcommand{\NN}{\ensuremath{\mathbb N}}
	\newcommand{\lintbraket}{\ensuremath{[\hspace{-1ex}{[}\hspace{0.5ex}{}}}
	\newcommand{\rintbraket}{\ensuremath{\hspace{0.5ex}{}]\hspace{-1ex}{]}\hspace{0.5ex}{}}}	
	\newcommand{\Argmin}[1]{\ensuremath{\mathrm{Arg}\underset{#1}{\mathrm{min}\,}}}
	
	\newcommand{\gaussian}[1]{\ensuremath{\mathcal{N}\left(#1\right)}}
	\newcommand{\uniform}[1]{\ensuremath{\mathcal{U}\left(#1\right)}}
	\newcommand{\vect}[1]{\boldsymbol{#1}}

	\newcommand{\iInt}[1]{\ensuremath{[\hspace{-0.1em}[ #1 ]\hspace{-0.1em}]}}

	\def\eg{e.g.,\ }
	\def\cf{\textit{cf.}\ }

	\newcommand{\train}[1]{\ensuremath{{#1}^{\mbox{\scriptsize Train}}}}
	\newcommand{\test}[1]{\ensuremath{{#1}^{\mbox{\scriptsize Test}}}}
	\newcommand{\val}[1]{\ensuremath{{#1}^{\mbox{\scriptsize Thr}}}}	
	\newcommand{\base}[1]{\ensuremath{{#1}^{\mbox{\scriptsize Base}}}}

	\newcommand{\bo}[1]{\textbf{#1}}	

\title{Feature Learning for Fault Detection in High-Dimensional Condition-Monitoring Signals}
\author{Gabriel Michau\\
ETH Zürich,\\
Swiss Federal Institute of Technology,\\
Zurich, Switzerland\\
\And 
Olga Fink,\\ 
ETH Zürich,\\
Swiss Federal Institute of Technology,\\
Zurich, Switzerland\\
}

\renewcommand{\today}{10th July 2019}

\begin{document}

\maketitle

\begin{abstract}
Complex industrial systems are continuously monitored by a large number of heterogeneous sensors. The diversity of their operating conditions and the possible fault types make it impossible to collect enough data for learning all the possible fault patterns. 
The paper proposes an integrated automatic unsupervised feature learning and one-class classification for fault detection that uses data on healthy conditions only for its training. The approach is based on stacked Extreme Learning Machines (namely Hierarchical, or HELM) and comprises an autoencoder, performing unsupervised feature learning, stacked with a one-class classifier monitoring the distance of the test data to the training healthy class, thereby assessing the health of the system.

This study provides a comprehensive evaluation of HELM fault detection capability compared to other machine learning approaches, such as stand-alone one-class classifiers (ELM and SVM), these same one-class classifiers combined with traditional dimensionality reduction methods (PCA) and a Deep Belief Network. The performance is first evaluated on a synthetic dataset that encompasses typical characteristics of condition monitoring data. Subsequently, the approach is evaluated on a real case study of a power plant fault.
The proposed algorithm for fault detection, combining feature learning with the one-class classifier, demonstrates a better performance, particularly in cases where condition monitoring data contain several non-informative signals.
\end{abstract}


\section{Introduction}
\label{sec:introduction}

Machine learning and artificial intelligence have recently achieved some major breakthroughs leading to significant progress in many domains, including industrial applications~\cite{bengio_representation_2013, Shao2017,  Valada2018, Li2018}. One of the major enablers has been the progress achieved on automatic feature learning, also known as representation learning~\cite{bengio_representation_2013}. It improves the performance of machine learning algorithms while limiting the need of human intervention.
Feature learning aims at transforming raw data into more meaningful representations simplifying the main learning task.

Traditionally, features were engineered manually and were thereby highly dependent on the experience and expertise of domain experts. This has limited the transferability, generalization ability and scalability of the developed machine learning applications~\cite{Khan2018}.
The feature engineering task is also becoming even more challenging as the size and complexity of data streams capturing the condition of industrial assets are growing.
These challenges have fostered the development of data-driven approaches with automatic feature learning, such as Deep Learning that allows for end-to-end learning~\cite{Khan2018}.

Feature learning techniques can be classified into supervised and unsupervised learning.
While in its traditional meaning, supervised learning requires examples of ``labels'' to be learned, supervised feature learning refers instead to cases where features are learned while performing a supervised learning task, as for example fault classification or remaining useful life (RUL) estimation in prognostics and health management (PHM) applications. Recently, deep neural networks have been used for such integrated supervised learning tasks, including Convolutional Neural Networks (CNN)~\cite{Ince2016, Krummenacher2018, Li2018b} or different types of recurrent neural networks, \eg Long-Short-Term-Memory (LSTM) networks  ~\cite{Zhao2017b}.
The performance achieved with such end-to-end feature learning architectures on commonly used PHM benchmark datasets was shown to be superior to traditional feature engineering applications~\cite{Li2018, Zhang2018}.

Contrary to the supervised learning, unsupervised feature learning does not require any additional information on the input data and aims at extracting relevant characteristics of the input data itself.
Examples of unsupervised feature learning include clustering and dictionary learning ~\cite{bengio_representation_2013}. Deep neural networks have also been used for unsupervised feature learning tasks for PHM applications, \eg with Deep Belief Networks ~\cite{Chen2017, Zhao2017}, and with different types of autoencoders~\cite{Chen2017}, \eg Variational Autoencoders \cite{Yoon2017}.

Even though the approaches described above applied either supervised or unsupervised feature learning within their applications, the subsequent step of using the learned features within fault detection, diagnostics or prognostics applications have always required labeled data to train the algorithms. Yet, while data collected by condition monitoring devices are certainly massive, they often span over a very small fraction of the equipment lifetime or operating conditions. In addition, when systems are complex, the potential fault types can be extremely numerous, of varied natures and consequences. In such cases, it is unrealistic to assume data availability on each and every fault that may occur, and these characteristics challenge the ability of data-driven approaches to actually learn the fault patterns without having access to a representative dataset of possible fault types.

The focus of the methodology proposed in this paper is to account for this possibility and to use healthy data only for the training. This is contrary to the widely used approaches applied for such problems:  defining the problem as a two-class classification task and trying to take the imbalance of the two classes (abundant healthy data and few faulty cases) into account. We propose to define this problem as a one-class classification task, designed to identify outliers to the class used during the training without any knowledge on the characteristics of the faulty system conditions. One-class classification comprises the steps of comparing the testing data with a reference dataset, here a healthy dataset, and the design of a decision boundary for outlier detection~\cite{Moya1996}. In this problem definition, we propose to interpret the raw output of the one-class classifier as a distance to the healthy class and therefore as a health indicator: the larger the deviation from the healthy system conditions, the larger the distance to the healthy class, and therefore the larger the value of the health indicator.  The decision boundary is a threshold on this distance, estimated thanks to healthy data not used in the training of the one-class classifier. This health monitoring concept is different to other approaches that proposed to use health indicators (HIs) to monitor the system condition~\cite{Malhotra2016}, as the HIs were learned in a supervised way, contrary to the proposed unsupervised learning approach. Thanks to the distance to the healthy class, the proposed approach can distinguish different degrees of fault severity.

Results presented in this paper demonstrate that in high dimension, the one-class classification is a challenging task and that results are improved by first learning the informative features. As healthy data only is used for the training, features have to be obtained in an unsupervised way and we propose to use a compressive autoencoder to perform this task. To the best of our knowledge, this is the first deep learning approach enabling efficient health monitoring trained solely with nominal condition monitoring data. This approach is unsupervised in the sense that it is used to detect degraded conditions that were not part of the training dataset.

The approach presented here takes advantage of the recent advances in using multi-layer extreme learning machines for representation learning~\cite{yang_autoencoder_2017}. The novelty here, is also in the use of hierarchical extreme learning machines (HELM)~\cite{cao_building_2016, michau_deep_2017, Michau_2018b}. HELM consists in stacking sparse autoencoders, whose neurons can directly be interpreted as features and a one-class classifier, aggregating the features in a single indicator, representing the health of the system.
The very good learning abilities of HELM have already been used in other fields, \eg in medical applications~\cite{miotto_deep_2016} and in PHM~\cite{michau_deep_2017}. This study provides a comprehensive evaluation of HELM detection capability compared to other machine learning approaches, including Deep Belief Networks.

The paper is organised as follows:
Section I details the HELM theory and the algorithms used for this research.
Section II presents a simulated case study designed to test and compare the HELM to other traditional PHM models. Controlled experiments allow for quantified results and comparisons. A real application is analysed in Section III, with data from a power plant experiencing a generator inter-turn failure.
Finally, results and perspectives are discussed in Section IV.

\section{Framework of Hierarchical Extreme Learning Machines (HELM)}
\label{sec:framework}

\begin{table}\small
\begin{center}
\setlength\tabcolsep{3pt}
\caption{Notations}
\label{tbl:notations}
\begin{adjustbox}{max width=\columnwidth}
\begin{tabular}{ll}
\toprule
\multicolumn{2}{l}{\textbf{Superscript Notations}} \\
\midrule
$\train{\cdot}$   & Training, healthy, dataset to train the neural networks                                    \\
$\val{\cdot}$     & Healthy dataset for finding the decision boundary (Eq.\eqref{eq:thrd})\\
$\test{\cdot}$    & Test data  \\
\midrule
\multicolumn{2}{l}{\textbf{Data Variables}}                                                                             \\
$X$               & Input data to the model                                                                    \\
$Y$               & Output values of a neural network                                                          \\
$T$               & Target data when training a neural network                                                 \\
$Z$               & Labels of datapoints                                                                       \\
\midrule
\multicolumn{2}{l}{\textbf{Dataset sizes}}   \\
$D$               & Input dataset dimension                                                                    \\
$D^Y$             & Dimension of the model output                                                              \\
$K$               & Number of sample in the dataset                                                            \\
\midrule
\multicolumn{2}{l}{\textbf{Neural Networks Variables}} \\
$L_i$             & Number of neurons in the i-th hidden layer of the HELM.                                    \\
$A$               & Weight matrix between input and hidden layer                                               \\
$B$               & Bias vector between input and hidden layer                                                 \\
$\beta$           & Weight Matrix between hidden layer and output                                              \\
$g$               & activation function                                                                        \\
$H$               & Hidden layer matrix: $\vect{H} = g(\vect{A}, X, B)$                                        \\
$C$               & Ridge regularization parameter in the one-class classifier                                 \\
$\lambda$         & Lasso regularization parameter in the autoencoder                                         \\
\midrule
\multicolumn{2}{l}{\textbf{Experiments}}  																				\\
$N_{\mbox{exp}}$ & Number of experiments                                                                      \\
TP                & True Positives:  \% of valid detections                                                     \\
FP                & False Positives: \% of false detections                                                     \\
TN                & True Negatives:  \% of non-detected healthy inputs                                          \\
FN                & False Negatives:  \% of missed detections                                                   \\
Acc               & Accuracy: (TP+TN)/2 																		\\
Mag				  & Magnification Coefficient                                                                    \\
$\gaussian{\mu,\sigma}$ & Gaussian distribution (average $\mu$, standard deviation $\sigma$)									\\
$\uniform{[a,b]}$ & Uniform distribution in the interval [a,b]													\\
$\uniform{\iInt{a,b}}$ & Uniform distribution of integers in [a,b]								\\
\bottomrule
\end{tabular}
\end{adjustbox}
\end{center}
\end{table}

\subsection{HELM for fault detection in brief}

\begin{figure}
	\centering
	\includegraphics[width=3.1in]{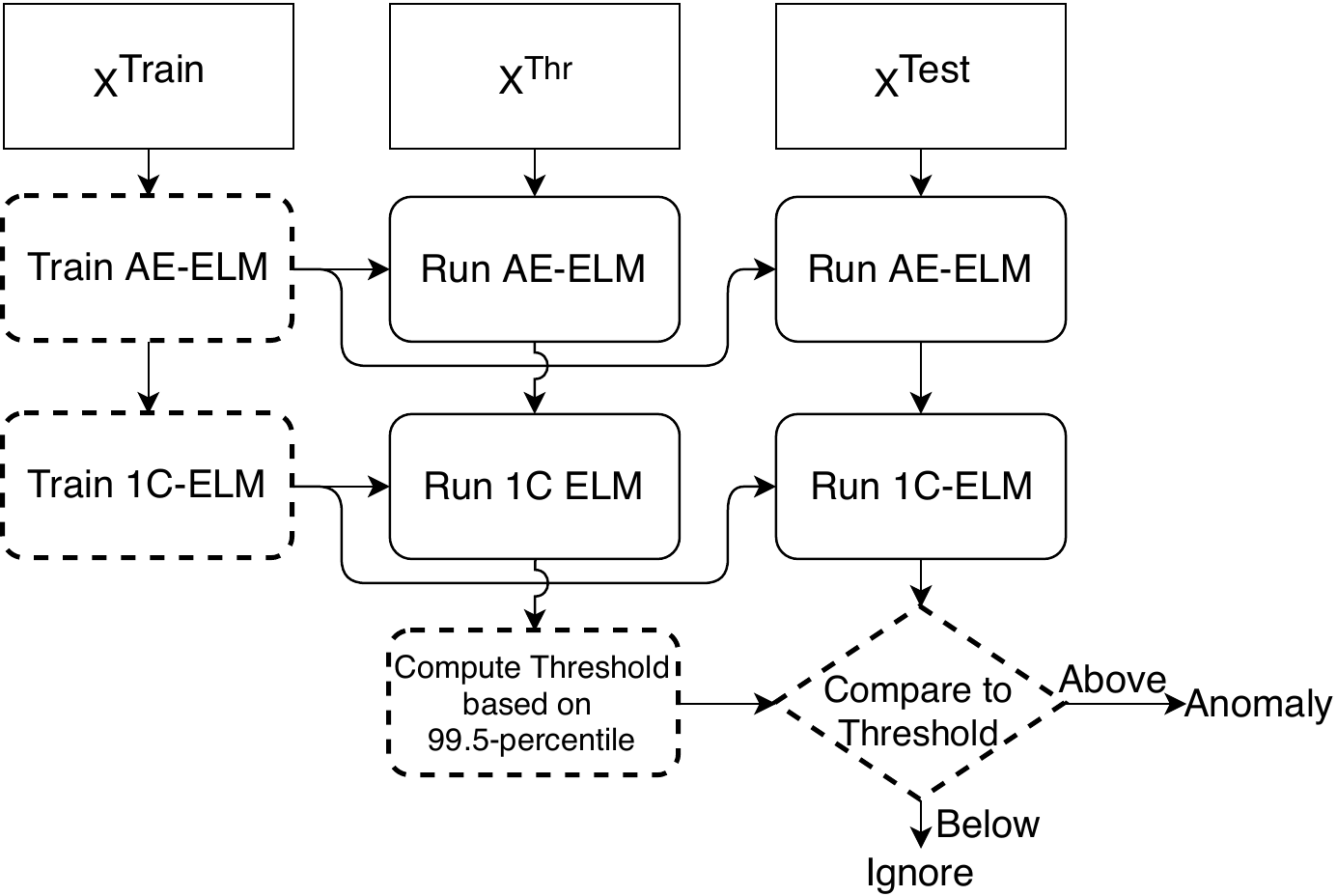}%
	\caption{HELM for Anomaly Detection Flow-Chart.}
	\label{fig:flowchart}
\end{figure}

\begin{figure}
	\centering
	\includegraphics[width=2.5in]{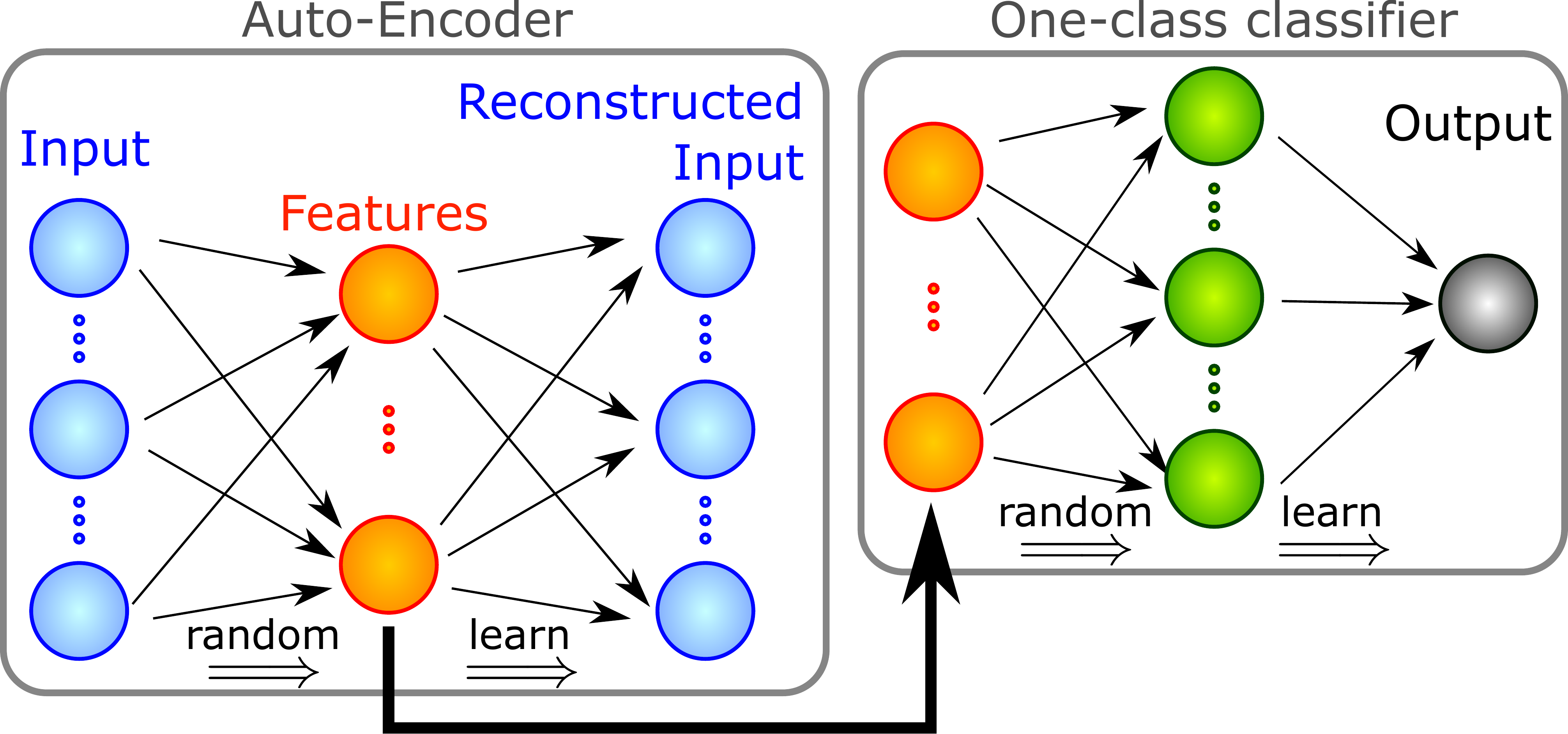}%
	\caption{HELM architecture: Example with a single AE whose the hidden layers is used as the input of the next layer: a one-class classifier.}
	\label{fig:HELM}
\end{figure}

HELM is a multilayer perceptron network, which was first proposed in~\cite{tang_extreme_2016}.
The structure of HELM consists in staking single layer neural networks hierarchically so that each one can be trained independently. In this paper, we propose to stack autoencoders and a one-class classifier. The compressive autoencoders are used for feature learning. The one-class classifier takes the features of the autoencoder as inputs and learns to output an indicator~\cite{michau_deep_2017} that can be interpreted as a health indicator.
Figure~\ref{fig:HELM} presents the proposed architecture, comprising a single autoencoder for feature learning and a one-class classifier. The framework proposed here is illustared in Figure~\ref{fig:flowchart} and can be simplified as:
\begin{itemize}
\item Split the training dataset (containing only observations from healthy system conditions) in two subsets: $\train{X}$, for training the neural network and $\val{X}$ for finding the optimal decision boundary of the one-class classification.
\item Train the autoencoder with $\train{X}$ such as to minimise the reconstruction error.
\item Train the one-class classifier network with the features of $\train{X}$ from the previous autoencoder such as to minimise the distance between the output and the healthy class represented by the value $1$.
\item Run both networks with $\val{X}$. Infer a minimal threshold on the output of the one-class classifier such that most points in $\val{X}$ belong to the healthy class.
\item When using testing data $\test{X}$, if the output of the one-class classifier is above threshold, consider these data as abnormal.
\item The value of the output of the one-class classifier provides an indication of the severity of the abnormality.
\end{itemize}

The remainder of this section presents the theoretical background of HELM in more detail and highlights the adaptations needed to match the specific requirements of PHM applications.

	\subsection{Theory and Training of ELM}
	\label{ssec:framework:theoretical}
Extreme Learning Machines (ELM) are Single-hidden Layer Feed-forward Neural networks (SLFNs)~\cite{huang_extreme_2004}

SLFN equation is usually written as
\begin{equation}
\label{eqn:elm}
Y = g(\vect{A}, X, B)\cdot \vect{\beta},
\end{equation}
where $X$ represents the input, $Y$ the output, $g$ the activation function, $\vect{A}$ and $\vect{\beta}$ are weights, and $B$ are the biases.

A mathematical proof~\cite{huang_universal_2006} demonstrates that given a linear kernel, $g$, a training set $X$ and its target $T$, a SLFN with enough neurons and randomly sampled $\vect{A}$ and $B$ can approximate $T$ with any accuracy. Such networks with randomly sampled weights and biases are called \emph{ELM}.
Formally, for any $\epsilon>0$, for $\sigma_1>0 ,u\in \NN$ and for enough neurons, $\vect{\hat{\beta}}$ exists such that:
\begin{equation}
\Vert g(\vect{A}\cdot X + B)\cdot \vect{\hat{\beta}} - T\Vert_u^{\sigma_1} < \epsilon
\label{eqn:elm_approx}
\end{equation}
\begin{equation}
\Vert \vect{H} \cdot \vect{\hat{\beta}} - T\Vert_u^{\sigma_1} < \epsilon\label{eqn:elm_approx_mat}
\end{equation}
where $\vect{H}$ denotes the hidden layer matrix $g(\vect{A}\cdot \train{X} + B)$.

By sampling randomly $\vect{A}$, the input weights, and $B$, the bias, training an ELM only consists in computing the optimal output weights $\vect{\beta}$. This simplifies Problem~\eqref{eqn:elm_approx_mat} considerably as it is now formulated as the single-variable regularised convex optimisation in Problem~\eqref{eqn:beta_est}.
\begin{equation}
\label{eqn:beta_est}
\vect{\hat{\beta}} = \Argmin{\vect{\beta}} \Vert \vect{H}\vect{\beta} - T \Vert_u^{\sigma_1} + C \Vert \vect{\beta} \Vert_v^{\sigma_2 }
\end{equation}
where $\sigma_1, \sigma_2 >0$, $u, v\in \NN\backslash 0$ and $C\geq 0$ is the weight of the regularization.
$C$ represents a compromise between the learning objective and properties of $\vect{\beta}$ that one would like to impose (\eg sparsity, non-diverging coefficient, etc...).
This is the strong advantage of ELM over regular SLFN. For traditional feed-forward neural networks, the weights $\vect{A}$, $\vect{\beta}$ and $B$ are optimised over several iterations, usually using a back-propagation algorithm, while for ELM, the optimisation of $\beta$ solely is a much faster and easier process and has an unique solution.

In Problem~\eqref{eqn:beta_est}, any combination of $\sigma_1$, $\sigma_2$, $u$, $v$ leads to a different solution of Problem~\eqref{eqn:beta_est}, it is however usual to take $(\sigma_1, \sigma_2, u, v)\in \lintbraket 1,2\rintbraket$.
When $\sigma_1 = \sigma_2 = 2$ and $u=v=2$, Eq.~\eqref{eqn:beta_est} is known as the ridge regression problem (also referred to as Tikhonov regularisation problem)~\cite{huang_what_2015}, and it has an explicit solution:
\begin{equation}
\label{eqn:ridge_sol}
\vect{\beta} = \left( C\cdot\II + \vect{H}^\top\vect{H}\right)^{-1}\vect{H}^\top T
\end{equation}

\subsection{ELM-Autoencoder for Feature Learning}
Autoencoders (or AE) are unsupervised machine learning tools trained for the reconstruction of their input. Structural or mathematical constraints make sure that they do not simply learn the identity operation, such as adding noise to the input (the autoencoder is then qualified as denoising), imposing dimensionality reduction (compressive AE), dimensionality increase or adding a kernel regression hidden layer (variational AE).

Within neural networks, autoencoders are often part of multilayer architectures: they can be trained independently to the problem at hand and are often used for feature learning~\cite{vincent_stacked_2010}. Conceptually, in a compressive AE, the hidden layers learn the best few combinations of the different measurements that can explain all the signals. The ELM framework can be used to train single layer autoencoder neural networks.
In this case, it consists in solving Problem~\eqref{eqn:beta_est} with the target $T$ equal to the input $X$.

One concern for practitioners is the possibility to understand how the signals are linked to the features and assess their impact on the results. This is a motivation to foster sparse connectivity between features and reconstructed inputs.  Intuitively, if each feature is connected to few measurements only, each feature should maximise the information it contains on only a part of the system. This would enable the monitoring engineers to track the changes in the features, by linking them to the parts of the system potentially affected.
Sparsity can be achieved using the $\ell_0$-norm for the regularization term ($v=0,\sigma_2 = 1$). Yet, this would make Problem~\eqref{eqn:beta_est} non-convex and require integer programming solver. A surrogate is the $\ell_1$-norm, which leads to the LASSO problem, one of the classic convex optimisation problem in the literature~\cite{Hastie_2017}.
One can note, that the $\ell_2$-norm has a closed form solution and is, thus, much easier to compute (\cf Eq.~\eqref{eqn:ridge_sol}). Yet, the solution is likely to be dense and redundant~\cite{cambria_extreme_2013}, which is not compatible with our idea of a good set of features.

The AE-ELM consists finally in solving the typical LASSO problem:
\begin{equation}
\label{eqn:beta_est_l1}
\vect{\hat{\beta}} = \Argmin{\vect{\beta}} \Vert \vect{H}\vect{\beta} - X \Vert_2^{2} + \lambda \Vert \vect{\beta} \Vert_1
\end{equation}
In this paper, we solve this problem using the FISTA algorithm (Fast Iterative Shrinkage-Thresholding Algorithm) known for both its fast convergence and computational simplicity~\cite{beck_fast_2009}. The process for solving Problem~\eqref{eqn:beta_est} is detailed in Algorithm~\eqref{alg:FISTA}.

\begin{algorithm}[!htbp]
	\caption{FISTA}
	\label{alg:FISTA}
	\begin{algorithmic}[1]
		\Input $\vect{H}, \quad X, \quad \lambda\geq 0,  \quad \delta\in ]0,1[, \quad \epsilon \geq 0 $
		\Let{$\gamma$}{$\frac{\delta}{1+\Vert \vect{H} \Vert_2}$}
		\State $k \gets 0, \quad \vect{\beta}_k \gets 0, \quad \vect{y}_k \gets 0,$
		\State $t_k \gets 1, \quad \mbox{Crit}\gets \infty,$
		\While{Crit\ $\geq \epsilon$}
			\State $\vect{c}_k = \vect{y}_{k} - 2 \cdot \gamma \vect{H}^\top \left(\vect{H} \vect{y}_{k} - X\right)$
			\State $\vect{\beta}_{k+1} = \max (\vert \vect{c}_k \vert - \lambda\cdot\gamma, 0)\cdot \mbox{sign}(\vect{c}_k)$
			\State $t_{k+1} = \frac{1}{2}\left( 1 + \sqrt{1+4 t_k^2}\right)$
			\State $y_{k+1} = \vect{\beta}_{k+1} + \frac{t_k - 1}{t_{k+1}} \left( \vect{\beta}_{k} -\vect{\beta}_{k+1} \right)$
			\State Crit\ $= \Vert \vect{\beta}_{k} -\vect{\beta}_{k+1} \Vert_2$
			\State $k = k+1$
		\EndWhile
		\Output $\vect{\beta}_{k}$
	\end{algorithmic}
\end{algorithm}
	
\subsection{Health Indicator and Magnification for Fault Detection}

In the context of critical and complex systems, the high number of possible faults due to the large number of parts involved in the system, the rarity of certain faults (if ever experienced at all), and, overall, the (understandable) lack of willingness from operators to let the faults happen repeatedly for data collection, make it impossible to aim for fault recognition or classification. Instead, we propose to focus on abnormality detection, training a one-class classifier ELM~\cite{leng_one-class_2015} on features from healthy data solely.

In this case, we use an ELM-based one-class classifier trained considering the ridge regression formulation in Eq.(5) with output $T=(1)_{1\leq k \leq K}$ ($D^Y = 1$). Then, the distance between the output and the normal class, represented by the value $1$, is monitored. A threshold on this distance discriminates between the normal and the faulty class. As such, this distance is similar to a health indicator~\cite{hu_deep_2016}. We propose here to base the threshold on another set $\val{X}$, that is, a set of datapoints from healthy conditions, not used in the training. This step ensures higher robustness of the threshold to variations in the healthy conditions and mitigates the potential risk of overfitting during training, which would result in a too small threshold, causing thereby too many false positives during the application. Experiments have shown that a good threshold can be designed as:
\begin{equation}
\label{eq:thrd}
\mbox{Thrd} = \gamma\cdot \mbox{percentile}_{p}(\vert 1 - \val{Y}\vert)
\end{equation}
where $\mbox{percentile}_p$ is the $p$th-percentile function with, $p,\gamma \geq0$ as hyperparameters.

Then the actual class $Z$ of a sample $i$ ($1$ for healthy, $-1$ otherwise) can be devised with the following equation later denoted as decision rule:
\begin{equation}
\label{eqn:decision_rule}
\test{Z}_i = \mbox{sgn}\left( \mbox{Thrd} - \vert 1 - \test{Y}_i\vert\right)
\end{equation}

The choice of both $p$ and $\gamma$ is to some extent a single problem, consisting in fact in choosing a robust threshold. In this paper, we chose to fix $p=99.5$ and experiments will show that $\gamma\in [1,3]$ provides good results.

We denote by \textit{magnification coefficient} the ratio between the distance of the one-class classifier output to 1 and the threshold:
\begin{equation}
\label{eq:mag}
\mbox{Mag} = \frac{(\vert 1 - \test{Y}\vert)}{\mbox{Thrd}}.
\end{equation}
This ratio is above 1 for points detected as abnormal and its value represents the robustness of the detection. A magnification coefficient well above the value 1 corresponds to a detection with low sensitivity to the choice of $\gamma$ and $p$ and therefore a higher confidence in the detection.

	\subsection{Stacking AE(s) and a one-class classifier for HELM}
A traditional approach is to pre-train an autoencoder before using it in a deeper classification architecture, which is subsequently trained and fine-tuned using back-propagation (BP)~\cite{lecun_deep_2015}. To the opposite, in HELM, each layer is trained sequentially and independently. The hidden layer of each successive ELM becomes the input of the next.
This avoids well-known limitations of the BP, in particular the risk of exploding or vanishing gradient and intensive computations required for the training~\cite{vincent_stacked_2010, hinton_fast_2006}.
In our proposed architecture of HELM, the lower layers are autoencoders, while the last layer is a one-class classifier, as illustrated in Figure~\ref{fig:HELM}.

If $N$ denotes the number of stacked autoencoders, then training the HELM corresponds to Algorithm~\ref{alg:HELMtrain}. Running the HELM, for threshold computation and testing, consists in applying Algorithm~\ref{alg:HELMrun}.
\begin{algorithm}
	\caption{Function HELM$^{\mbox{Train}}$ (Training of a HELM)}
	\label{alg:HELMtrain}
	\begin{algorithmic}[1]
		\Input $C \geq 0, \quad \lambda \geq 0, \quad N\in \NN \quad \train{X}, \quad \train{T}$
		\Let{$\vect{x}_1$}{$\train{X}$}
		\For{$i=1, \ldots, N$} \Comment{\textit{Stacked AE ELM}}
			\State \textbf{Generate:} $\vect{A}_i, B_{i}$, random weights
			\State $\vect{H}_i = g(\vect{x}_i, \vect{A}_i, B_i)$
			\State $\vect{\beta}_i = \Argmin{\vect{\beta}} \lambda \Vert \vect{\beta} \Vert_1 + \Vert \vect{H}_i\vect{\beta} - \vect{x}_i \Vert_2^{2}$ {\footnotesize\Comment{(\cf Alg.~\ref{alg:FISTA})}}
			\State $\vect{x}_{i+1} = \vect{x}_i \cdot \vect{\beta}_i^\top$
		\EndFor
		\Comment{\textit{Upper layer ELM}}
		\State \textbf{Generate:} $\vect{A}_{N+1}, B_{N+1}$, random weights
		\State $\vect{H}_{N+1} = g(\vect{x}_{N+1}, \vect{A}_{N+1}, B_{N+1})$
		\State $\vect{\beta}_{N+1} = \Argmin{\vect{\beta}} C \Vert \vect{\beta} \Vert_2^{2} + \Vert \vect{H}_{N+1}\vect{\beta} - \train{T} \Vert_2^{2}$
		\Output $\lbrace \vect{\beta}_i \rbrace_{\tiny 1\leq i \leq N+1}, \quad \left(\vect{A}_{N+1}, B_{N+1}\right), g$
	\end{algorithmic}
\end{algorithm}

\begin{algorithm} [!htbp]
	\caption{Function HELM$^{\mbox{Run}}$ (Running a HELM)}
	\label{alg:HELMrun}
	\begin{algorithmic}[1]
		\Input $X, \quad \mbox{HELM}^{\mbox{Train}}(C, \  \lambda, \  N, \  \train{X}, \  \train{T})$
		\Let{$\vect{x}_1$}{$X$}
		\For{$i=1, \ldots, N$}
			\State $\vect{x}_{i+1} = \vect{x}_i \cdot \vect{\beta}_i^\top$
		\EndFor
		\State $\vect{H} = g(\vect{x}_{N+1}, \vect{A}_{N+1}, B_{N+1})$
		\Output $Y = \vect{H}\vect{\beta}_{N+1}$
	\end{algorithmic}
\end{algorithm}

In the following, HELM with a single AE is tested on a simulation and on a real case study. HELM is compared to five other approaches:
First, to assess the benefits of feature learning, HELM is compared to two one-class classifiers (ELM and Support Vector Machine, SVM). Second, features are devised through one of the most traditional ways, a PCA, and are then used as input to two one-class classifiers (ELM and SVM).
Last, to mimic the HELM architecture with another commonly used neural network, HELM is compared with a Deep Belief Network (DBN) composed of two staked RBM: one for feature learning and a second whose hidden layer is fully connected to a single output neuron trained as a one-class classifier.

\noindent \textbf{Remark about the one-class classifier SVM:}
Note that usually, the one-class classifier SVM output is interpreted as the likelihood of a point to belong to the main class, under two assumptions: First, that the data follows the distribution used as a kernel (Gaussian most of the times), second, that the number of outliers in the training data was correctly specified. In this traditional interpretation, $0$ is the decision boundary and points with positive likelihood are labeled as belonging to the training class (healthy), while points with negative likelihood are labeled as outliers. The quality of the classification depends therefore strongly on the validity of the two above assumptions. In our experiments, better detection performances could be achieved by choosing a decision boundary on the negative likelihood with Eq.~\eqref{eq:thrd} instead of $0$. This modification compared to the traditional implementation provides a fairer comparison to other approaches.

\section{Simulated Case Study}
\label{sec:simulated_case}
To evaluate the performances of the proposed approaches under known and controlled conditions, a simulated case study is designed. The synthetically generated datasets aim to represent behaviors observed in real condition-monitoring signals and to simulate varied faults.
All the following experiments have been performed with Matlab on a Laptop (Intel Core i7-4600M 2.90GhZ, 16 GB of DDR3 ram).

	\subsection{Description}
	\label{ssec:sc:desc}
\noindent \textbf{Methodology:}

\begin{algorithm}\footnotesize
	\caption{Simulated Case Study}
	\label{alg:SimCS}
	\begin{algorithmic}[1]
	\algrenewcommand\algorithmicindent{1em}
		\Let{$K$}{$14000$ samples}
		\Let{$D$}{$200$ sensors}
		\For{$n\in[5,10]$, $\quad f\in [Id(),log()]$}
			\MRepeat \ \ $N_{\mbox{exp}} = 150$ times
			\State \textbf{Draw}: $\base{\vect{\alpha}}  \in \left[\gaussian{0,1}\right]^{n}$, $\alpha_f  \in \gaussian{0,1}$
			\State \textbf{Draw}: $\base{\vect{\epsilon}} \in \left[\uniform{[0,3]}\right]^{n}$, $\epsilon_f \in \uniform{[0,3]}$
			\State \textbf{Draw}: $\base{X} \in \left[\gaussian{0,1}\right]^{n\times K}$
			\State \textbf{Draw}: $X_f \in \left[\gaussian{0,1}\right]^{1\times 1000}$
			\Let{$\base{X}$}{$\base{\vect{\alpha}}\cdot \base{X} + \base{\vect{\epsilon}}$}
			\State{$\base{X}[9000:10000,1]\ \ \ast\hspace{-0.4em}=\ \  1.2$}
			\State{$\base{X}[10000:11000,1]\ \ \ast\hspace{-0.4em}=\ \ 1.5$}
			\State{$\base{X}[11000:12000,1]\ \ +\hspace{-0.4em}=\ \ 0.2\cdot \base{\vect{\alpha}}[0]$}
			\Let{$\base{X}[12000:13000,1]$}{$\alpha_f\cdot X_f + \epsilon_f$}
			\State \textbf{Draw}: $\vect{\alpha}\in \left[\uniform{[0,1]}\right]^{D}$
			\State \textbf{Draw}: $\vect{b}\in \left[\uniform{\iInt{1,n}}\right]^{D}$
			\State \textbf{Draw}: $\vect{\epsilon}\in \gaussian{0,0.01\cdot f(\vect{\alpha}\cdot \base{\vect{\alpha}}[\vect{b}]) }$
			\Let{$X$}{$f\left(\vect{\alpha}\cdot \base{X}[:,\vect{b}]\right) + \vect{\epsilon}$}
			\State \textbf{Draw}: $\vect{i_f}\in \left[\uniform{\iInt{1,D}}\right]^{10}$
			\State{$X[13000:14000,\vect{i_f}] \ \ \ast\hspace{-0.4em}=\ \  1.2$}
			\For{Every \textit{Model} and every set of hyperparameters}
				\State{Train \textit{Model} on $X[0:7000,:]$}
				\State{\textbf{Compute} $\mbox{Thrd}$ in Eq.\eqref{eq:thrd} with $X[7000:8000,:]$}
				\State{\textbf{Compute} FP on $X[8000:9000,:]$}
				\State{\textbf{Compute} TP on each fault type (fault \textit{1} on $X[9000:10000,:]$, ..., fault \textit{5} on $X[13000:14000,:]$)}
				\State{\textbf{Compute} Mag, the average over true positives of the output value of the one-calls classifier, normalised by $\mbox{Thrd}$.}
				\EndFor
			\EndRepeat
		\EndFor
		\State \textbf{Average} TN over $N_{\mbox{exp}}$ for each \textit{Model}, hyperparameter set, value $n$ and function $f$.
		\State \textbf{Average} TP and Mag over $N_{\mbox{exp}}$ for each \textit{Model}, hyperparameter set, value $n$, function $f$ and fault type.
		\State \textbf{Compute} Prec$=\frac{TP}{TP+FP}$ for each model, hypeparameter set, $n$ value, function $f$ and fault type.
	\end{algorithmic}
\end{algorithm}

We generate $150$ experiments according to Algorithm~\ref{alg:SimCS}. The core of the experiment is to simulate measurements from $D=200$ sensors, on a system with $n$ intrinsic physical properties (\eg temperature, rotation, voltage etc...). Each of the $D=200$ sensors, denoted $s$, is reading one of the $n$ intrinsic properties according to an internal reading function $f$, a sensitivity $\alpha_s$ and a noise $\epsilon_s$.
At a given time, we impose one of the following 5 faults on the measurements:
\begin{enumerate}
	\item One intrinsic property is multiplied by 1.2
	\item One intrinsic property is multiplied by 1.5
	\item A constant value computed as 20\% of the signal amplitude is added to one intrinsic property
	\item The intrinsic property changes
	\item The measurements of 10 sensors are multiplied by 1.2.
\end{enumerate}

\noindent \textbf{Choice of the Metrics:}
For each experiment, we generate one \textit{negative} set (healthy dataset) and one \textit{positive} set for each fault. In the context of one class-classification, we have the following relationships: positive (faulty) sets are either detected (true positive sets TP) or missed (false negative sets FN). Similarly, negative sets are either not detected as faulty (true negative sets TN) or falsely detected (false positive sets FP). Therefore, we have 
\begin{equation}
    N = TP+FN = TN+FP
\end{equation}
where $N$ is the number of experiments on which the metrics are computed ($N=50$ for the hyper-parameter tuning and $N=100$ for the reported results). The typical performance metrics are thus computed as follows:
\begin{itemize}
\item The true positive rate: $TPR = TP/N$
\item The true negative rate: $TNR = TN/N$
\item The false positive rate: $FPR = FP/N = 1-TNR$
\item The false negative rate: $FNR = FN/N = 1-TPR$
\item The precision is $TP/(TP+FP)$
\item The accuracy is $(TP+TN)/(2N)$
\item The $f1$-score is $2TP/(N+FP+TP)$
\end{itemize}

In the context of one class classification, the practitioners have two concerns. The first one is the minimisation of false alarm occurrences. In practical applications, verifying the occurred alarms takes time, expertise and is often associated with high costs. Too many of them will make the model impracticable to use. Second, as abnormalities are unexpected, random, rare and might have large consequences, the probability of raising an alarm in case of abnormal behaviour should be as high as possible. The two corresponding indicators are the FPR and the TPR. This is different to a two-class classification problem with imbalanced classes, where the $f1$ score would be favoured, reflecting that the classification of both classes is equivalently important and that the results need to account for possible class imbalance.

In the following, we report the results with the metrics TPR and FPR. An optimal model would at the same time minimise the FPR while maximising the TPR. Considering that TNR=1-FPR, we choose to report the results for the models maximising the average between the TPR and the TNR, that is, the accuracy. Depending on the priority of practitioners, one or the other could be weighted more, or other metrics could be used.
In such a one-class classification problem, most of the commonly used metrics can be computed with the knowledge of the TPR and of the FPR.
We report in addition the Magnification coefficient (Mag), as an indicator of the confidence in the fault detection.

\noindent \textbf{Model hyperparameters:}
The hyperparameter search is done according the following grid:
\begin{itemize}
\item $\gamma \in [1.1, 1.2, 1.5, 1.7, 2, 2.5, 3]$, the factor used in the decision rule in Eq.~\eqref{eq:thrd}
\item $L_{1} \in [5, 10, 20, 40, 70, 100]$, the number of neurons for the autoencoder (or DBN 1st layer)
\item $L_{2}\in [20, 50, 100, 200, 400, 800]$, the number of neurons for the one-class classifier (or DBN 2nd layer)
\item $\lambda \in [10^{-5}, 10^{-3}, 10^{-2}, 10^{-1}, 1]$, the LASSO regularisation parameter
\item $C \in [10^{-5}, 10^{-3}, 10^{-2}, 10^{-1}, 1]$, the ridge regularisation parameter
\item $L_{\tiny PCA} \in [5, 10, 15]$, the number of principle components selected in the PCA (up to 99\% of explained variance).
\item $S \in [10^{-5}, 10^{-1}, 1, 5, 10]$ the SVM Gaussian kernel scale
\item $\%_{\mbox{Out}} \in [0.5\%, 1\%, 2\%, 5\%]$ the number of outliers allowed in the training of SVM.
\end{itemize}

\noindent \textbf{Simulation parameters:}
Each of the $n$ intrinsic properties is simulated with a distinct random ``base'' signals $X^{\mbox{Base}}$. We report results for $n=5$ and $n=10$.
The sensor reading function $f$ is either the identity (linear measurement of the physical property) or the logarithm (often the case in intensity measurement). The measurements have an additive random Gaussian-noise $\epsilon_s$, corresponding to $1\%$ of the reading amplitude.

For each experiment, $K=14000$ datapoints are generated. $7000$ are used for training the models, $1000$ are used for computing the decision threshold in Eq.\eqref{eq:thrd} and another $1000$ are used to compute the false positives ratio (FPR). The remaining $5000$ datapoints are divided in 5 subgroups of size $1000$, each subgroup being impacted by one the the 5 faults. The model is tested on each faulty set to compute the TPR per fault type.
The statistics are aggregated over multiple repetitions of the experiment. We first repeat the experiments $50$ times to find the best set of hyperparameters with respect to the accuracies, we then report aggregated results on further $100$ simulations.

\noindent \textbf{Impact of randomness on (H)ELM efficiency:}
Prior experiments to this study demonstrated that for ELM-based methods, averaging the results over 5 runs with independent training reduces deviations in accuracy without changing its average. This is important for practitioners, as HELM is inexpensive to train and run and as reducing the randomness impact of the random weights and biases ($\vect{A}$ and $B$) on the results is a valid concern.

	\subsection{Results}

First, to demonstrate the ability of each model to detect the different faults and to give the model performance upper limits, the hyperparameters are optimised in order to maximise the accuracy on each fault independently over the first $50$ experiments. Results on the following $100$ experiments are reported in Table~\ref{tbl:cs-rslt} both for $n=5$, $n=10$ and for both reading function $f$.

\begin{figure}
	\centering
	\includegraphics[width=3.2in]{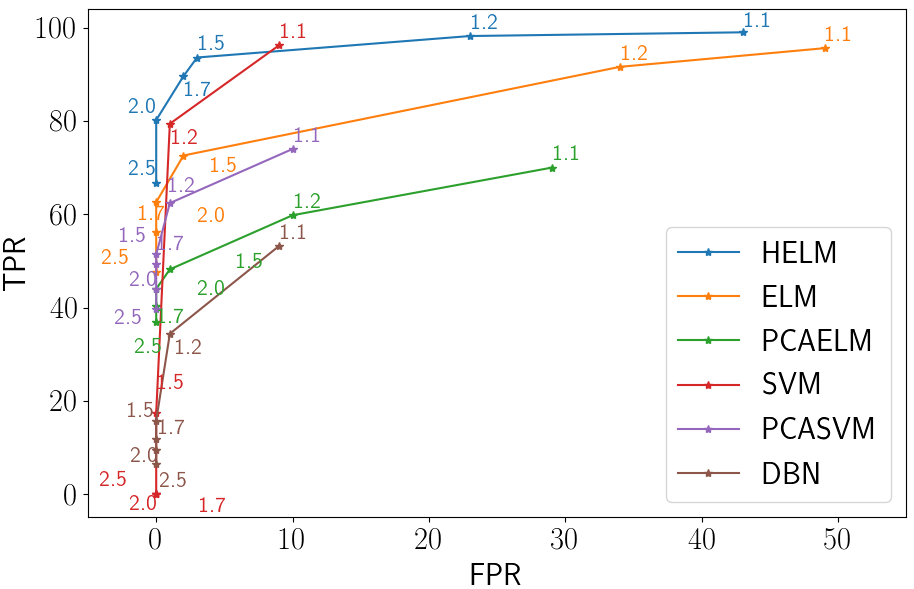}%
	\caption{ROC curve ($f=\log$ \& $n=5$). Each point of each curve is labelled with the corresponding value of $\gamma$, increasing from right to left.}
	\label{fig:ROC}
\end{figure}

Second, to demonstrate the robustness of the model to detect, with a single training, different kind of faults, the hyperparameters are tuned over the first $50$ experiments in order to maximise the average accuracy over the 5 faults.
Results aggregated over the following $100$ experiments and selected hyperparameters are reported in Table~\ref{tbl:cs-rslt_gen} and the sensitivity to the threshold $\gamma$ is presented as a ROC curve in Figure~\ref{fig:ROC}.

In both tables, for each column, the best results are presented in bold.
\begin{table*}[t] \small
	\begin{center}
\setlength\tabcolsep{3pt}
\caption{Performance of the different models, $n=5$ and $n=10$ base signals, for fault types \textit{1} to \textit{5}. Each cell contains three values: Accuracy (TPR/FPR).}
\label{tbl:cs-rslt}
$X = \vect{\alpha}\cdot \base{X}[:,\vect{b}] + \vect{\epsilon}$\\[0.1em]
\begin{adjustbox}{max width=\textwidth}
\begin{tabular}{lcccccccccccccc}
\toprule
Fault type      & \multicolumn{2}{c}{\textit{1}} &  & \multicolumn{2}{c}{\textit{2}}   &  & \multicolumn{2}{c}{\textit{3}}     &  & \multicolumn{2}{c}{\textit{4}}     &  & \multicolumn{2}{c}{\textit{5}}         \\ \cmidrule{2-3}\cmidrule{5-6}\cmidrule{8-9}\cmidrule{11-12}\cmidrule{14-15}
n       & 5       & 10      & & 5        & 10      & & 5       & 10      & & 5       & 10      & &  5    & 10 \\
\midrule
HELM   & \textbf{96 (96/4)} & \textbf{75 (64/14)} &  & \textbf{100 (100/0)} & \textbf{99 (99/1)} &  & \textbf{95 (96/6)} & \textbf{77 (78/24)} &  & \textbf{94 (88/0)} & \textbf{94 (87/0)} &  & 100 (100/1) & 100 (100/0) \\
ELM    & 84 (79/12)         & 64 (57/30)          &  & 99 (98/0)            & 90 (95/15)         &  & 85 (81/12)         & 66 (61/30)          &  & 93 (86/0)          & 91 (83/1)          &  & 99 (98/0)   & 98 (97/0)   \\
PCAELM & 64 (57/30)         & 63 (49/23)          &  & 88 (83/8)            & 85 (93/23)         &  & 69 (67/30)         & 66 (54/23)          &  & 91 (85/3)          & 91 (86/5)          &  & 100 (100/0) & 100 (100/0) \\
SVM    & 86 (90/19)         & 60 (21/2)           &  & \textbf{100 (100/0)} & 98 (97/2)          &  & 88 (94/19)         & 64 (30/2)           &  & 93 (85/0)          & 92 (86/2)          &  & 98 (96/0)   & 98 (97/2)   \\
PCASVM & 69 (62/24)         & 64 (47/20)          &  & 98 (97/1)            & 94 (91/4)          &  & 76 (77/25)         & 67 (53/20)          &  & 93 (87/1)          & 90 (84/4)          &  & 100 (100/0) & 100 (100/0) \\
DBN    & 70 (67/27)         & 67 (59/25)          &  & 84 (81/14)           & 80 (67/8)          &  & 67 (61/27)         & 65 (55/25)          &  & 84 (68/1)          & 85 (78/8)          &  & 87 (74/0)   & 89 (86/8)  \\

\bottomrule
\end{tabular}
\end{adjustbox}
\\[0.6em]
$X = \log(\vect{\alpha}\cdot \base{X}[:,\vect{b}]) + \vect{\epsilon}$\\[0.1em]
\begin{adjustbox}{max width=\textwidth}
\begin{tabular}{lcccccccccccccc}
\toprule
Fault type      & \multicolumn{2}{c}{\textit{1}} &  & \multicolumn{2}{c}{\textit{2}}   &  & \multicolumn{2}{c}{\textit{3}}     &  & \multicolumn{2}{c}{\textit{4}}     &  & \multicolumn{2}{c}{\textit{5}}         \\ \cmidrule{2-3}\cmidrule{5-6}\cmidrule{8-9}\cmidrule{11-12}\cmidrule{14-15}
n       & 5       & 10      & & 5        & 10      & & 5       & 10      & & 5       & 10      & &  5    & 10 \\
\midrule
HELM   & \textbf{93 (91/6)} & \textbf{73 (76/30)} &  & \textbf{100 (100/0)} & 97 (97/3)          &  & \textbf{94 (91/3)} & \textbf{77 (74/20)} &  & \textbf{96 (91/0)} & \textbf{94 (87/0)} &  & 100 (100/0) & 100 (100/0) \\
ELM    & 73 (79/34)         & 67 (67/34)          &  & 99 (99/2)            & 92 (95/12)         &  & 76 (86/34)         & 69 (50/12)          &  & 94 (89/2)          & 92 (84/0)          &  & 100 (100/0) & 100 (100/0) \\
PCAELM & 58 (49/33)         & 64 (59/32)          &  & 84 (81/13)           & 78 (81/25)         &  & 58 (47/31)         & 61 (53/32)          &  & 92 (84/1)          & 89 (78/0)          &  & 100 (100/0) & 100 (100/0) \\
SVM    & 92 (93/9)          & 59 (19/2)           &  & 100 (100/1) & \textbf{98 (98/2)} &  & 93 (94/9)          & 60 (22/2)           &  & 95 (90/1)          & 93 (88/2)          &  & 100 (100/0) & 100 (100/0) \\
PCASVM & 66 (42/10)         & 58 (29/14)          &  & 98 (97/1)            & 90 (93/13)         &  & 65 (39/9)          & 61 (35/14)          &  & 94 (89/1)          & 91 (82/0)          &  & 100 (100/0) & 100 (100/0) \\
DBN    & 65 (67/37)         & 62 (54/30)          &  & 78 (67/11)           & 71 (72/30)         &  & 64 (37/9)          & 60 (35/16)          &  & 82 (65/1)          & 84 (69/1)          &  & 90 (82/2)   & 92 (84/1) \\
\bottomrule
\end{tabular}
\end{adjustbox}
\end{center}
\end{table*}

From Table~\ref{tbl:cs-rslt}, two main results can be observed. First, according to expectations, accuracies are decreasing for higher signal complexity (increasing number $n$ of base signals and logarithmic measurements). Second, HELM achieves very good detection performances on all fault types and is the most efficient approach to detect faults that impact the base signals. HELM provides consistent results both for the different number of affected base signals and also for the two different measurement functions.

The benefits from the feature encoding layer is proven by the highest performance of HELM over the standard ELM. The robustness of HELM is also proven when comparing to the results of the one-class classifier SVM. While SVM performs equally well as HELM for $n=5$, its accuracy is strongly reduced when the inherent dimensionality of the measurements increases to $n=10$.
Using PCA as a first step for dimensionality reduction did not overall improve the results compared to using the one-class classifiers alone. Only on fault \textit{5}, PCA tends to improve the results, yet not significantly. This fault corresponds to the sensors being directly impacted, and thus would have the most direct impact on the principal component. Otherwise, the non-linear relationships between measurements and faults seem to limit the efficiency and impact of PCA.
Last, the Deep Belief Network shows robust results for $n=5$ or $n=10$, but with low accuracies.

Table~\ref{tbl:cs-rslt_gen} presents the accuracies and magnification coefficient for the set of hyperparameters with best average accuracy $\widehat{\mbox{Acc}}$ over the 5 fault types, the corresponding set of hyperparameter and the training time. Results discussed sofar are confirmed, in particular that HELM is the most robust model: with a single set of hyperparameters, it achieves almost optimal results on all faults, with accuracies consistently around 95\% and very strong magnification coefficients (up to around 100).
SVM is up to 100 times slower to train and its magnification is very low. Performing PCA before the SVM reduces the time needed for training and increases the magnification, but lowers the average accuracy.

The sensitivity analysis on Figure~\ref{fig:ROC} confirms that SVM, with a low magnification coefficient, has strongly decreasing performances as $gamma$ increases. For HELM, $\gamma\in[1.5-2.5]$ outperforms most of the other model settings.

\begin{table*}[t] \small
\setlength\tabcolsep{2pt}
\centering
\caption{Accuracies and Magnification for the set of hyperparameters maximising $\widehat{\mbox{Acc}}$ for fault types \textit{1} to \textit{5}.}
\label{tbl:cs-rslt_gen}
$X = \log(\vect{\alpha}\cdot \base{X}[:,\vect{b}]) + \vect{\epsilon}$ \quad $n=5$\\[0.1em]
\begin{adjustbox}{max width=\textwidth}
\begin{tabular}{llllllllllllllll|l|llllllll|l}
\toprule
& & \multicolumn{2}{c}{\textit{1}} & & \multicolumn{2}{c}{\textit{2}} & & \multicolumn{2}{c}{\textit{3}} & & \multicolumn{2}{c}{\textit{4}} & & \multicolumn{2}{c|}{\textit{5}} &  &             \multicolumn{8}{c|}{Hyperparameters} &     Time (s)     \\
\cmidrule{3-4}\cmidrule{6-7}\cmidrule{9-10}\cmidrule{12-13}\cmidrule{15-16}\cmidrule{18-26}
       & $\widehat{\mbox{Acc}}$ & Acc (TP/FP)        & Mag          &  & Acc (TP/FP)         & Mag          &  & Acc (TP/FP)        & Mag          &  & Acc (TP/FP)        & Mag         &  & Acc (TP/FP) & Mag          && $\gamma$ & $L_{1}$ & $L_{2}$ & $\lambda$ & $C$       & $L_{PCA}$ & $\%_{\mbox{Out}}$ & $S$ & \\
\midrule
HELM   & \textbf{95} & \textbf{91 (85/3)} & \textbf{2.3} &  & \textbf{99 (100/3)} & \textbf{6.3} &  & \textbf{94 (91/3)} & \textbf{2.2} &  & \textbf{95 (92/3)} & \textbf{88} &  & 99 (100/3)  & \textbf{105} && 1.5      & 20      & 100     & $10^{-2}$ & $10^{-5}$ &           &                   &     & 0.16     \\
ELM    & 85          & 68 (37/2)          & 1.4         &  & 95 (95/4)           & 1.9         &  & 68 (38/2)          & 1.4         &  & 89 (89/12)         & 5.5         &  & 99 (100/2)  & 6            && 1.5      &         & 400     &           & $10^{-5}$ &           &                   &     & 0.1      \\
PCAELM & 75          & 56 (22/10)         & 1.8         &  & 81 (72/10)          & 2         &  & 54 (17/10)         & 1.7         &  & 89 (88/10)         & 28          &  & 95 (100/10) & 70           && 1.5      &         & 100     &           &           & 15        &                   &     & 0.2      \\
SVM    & 94          & 91 (92/9)          & 1.2         &  & 96 (100/9)          & 1.3         &  & 93 (94/9)          & 1.2         &  & 93 (94/9)          & 1.7        &  & 96 (100/9)  & 1.5         && 1.1      &         &         &           &           &           & 1                 & 10  & 10.5     \\
PCASVM & 82          & 66 (42/10)         & 1.3          &  & 94 (98/10)          & 1.5         &  & 65 (40/10)         & 1.4         &  & 90 (90/10)         & 13          &  & 95 (100/10) & 28           && 1.2      &         &         &           &           & 15        & 1                 & 10  & 3.3      \\
DBN    & 72          & 63 (35/9)          & 1.2          &  & 76 (61/9)           & 1.3         &  & 62 (32/9)          & 1.1         &  & 81 (70/9)          & 1.8        &  & 80 (68/9)   & 2.0          && 1.1      & 70      & 100     &           &           &           &                   &     & 8.2  \\
\bottomrule
\end{tabular}
\end{adjustbox}
\end{table*}

\section{Real-Application Case Study: Generator Fault Detection}
\label{sec:generator_case}

	\subsection{Power Plant Condition Monitoring}
For the real case study, an H$_2$-cooled generator from an electricity production plant is evaluated. The generator is working on nominal load, with changing operating conditions and thus, variations in the parameters.
The evaluated dataset cover nine months (275 days) with 310 sensors recording every 5 minutes ($K=55\,774 $ and $D=310$). Sensors are grouped in 5 families:
\begin{enumerate}
\item The rotor shaft voltage is used to detect shaft grounding problems, shaft rubbing, electro-erosion, bearing isolation problems and rotor inter-turn shorts.
\item Rotor flux is used to detect the occurrence, the magnitude and the location of the rotor winding inter-turn short circuit.
\item Partial discharge is used to detect ageing of the main insulation, loose bars or contact as well as contamination.
\item End winding vibration is mainly used to detect deterioration in mechanical stiffness of the overhang support system.
\item Last, the Stator bar water temperature is also monitored.
\end{enumerate}

The generator failure mode tested here has been \textit {a posteriori} explained by expert as, first, at day 169, an intermittent short circuits in the rotor that remained undetected, denoted in the following by \textit{lower level fault}, developing in a second faulty state or \textit{upper level fault} at day 247, when the short circuit worsened to a continuous one. This led to the power plant shut-down.

	\subsection{Methodology}
Similarly as in the previous case study, the data are split in three, for training and threshold computation (first 120 days with a random sampling $94\%/6\%$) and for testing (from day 120 to day 169 to compute FP, the rest to compute TP). The dataset is hence split as follows: $\train{K} = 22\,017$, $\val{K} = 1\,406$ and $\test{K} = 4\,524 + 27\,827 $. Remind that $D=310$. As results need to be assessed on a single experiment, in the present case study, the TP and FP are the proportion of points for which the model output exceeds the threshold \emph{after} and \emph{before} day 169 respectively.

	\subsection{Results}

\begin{table}[t] \small
	\centering
	\caption{Performances on Real Dataset}
	\label{tbl:rslt_detection_real}
	\setlength\tabcolsep{3pt}
	\begin{adjustbox}{max width=\columnwidth}
	\begin{tabular}{lllll|c}
	\toprule
	 		& Acc      & TP (\%)    & FP (\%) & Mag     & Time (s) \\\midrule
	HELM    & \bo{95} & \bo{89.1} & 0.0     & 20      & 0.5 \\
	ELM     & 63      & 26.6      & 1.0     & 4.8     & 0.4      \\
	PCA ELM & 57      & 14.6      & 0.5     & 3.1     & 1.4  \\
	SVM     & 87      & 73.6      & 0.0     & 1.04    & 169     \\
	PCA SVM & 69      & 39.8      & 1.2     & 1.03    & 41 \\
	DBN     & 78      & 56.2      & 0.0     & \bo{40} & 5.5   \\\bottomrule
	\end{tabular}
	\end{adjustbox}
\\[0.6em]
Hyperparameters\\[0.1em]
\setlength\tabcolsep{3pt}
\begin{adjustbox}{max width=\columnwidth}
\begin{tabular}{lcccccccc}
\toprule
\       & $\gamma$ & $L_{1}$ & $L_{2}$ & $\lambda$ & $C$  & $L_{PCA}$ & $\%_{\mbox{Out}}$ & $S$ \\
\midrule
HELM    & 1.2      & 20      & 200     & $10^{-3}$ & $10^{-5}$ &           & 	   &     \\
ELM     & 1.2      &         & 200     &           & $10^{-5}$ &           & 	   &     \\
PCA ELM & 1.1      &         & 400     &           &      & 15        & 	   &     \\
SVM     & 1.1      &         &         &           &      &           &    1   & 5  \\
PCA SVM & 1.1      &         &         &           &      & 15        &    1   & 1  \\
DBN     & 1.1      & 10      & 50      &           &      &           &        &     \\
\bottomrule
\end{tabular}
\end{adjustbox}
\end{table}

\begin{figure*}
	\captionsetup[subfloat]{farskip=2pt,captionskip=5pt}
	\centering
	\subfloat[HELM (TP: 89.1\% \--- FP: 0.0\%)]
	{\includegraphics[width=8.2cm]{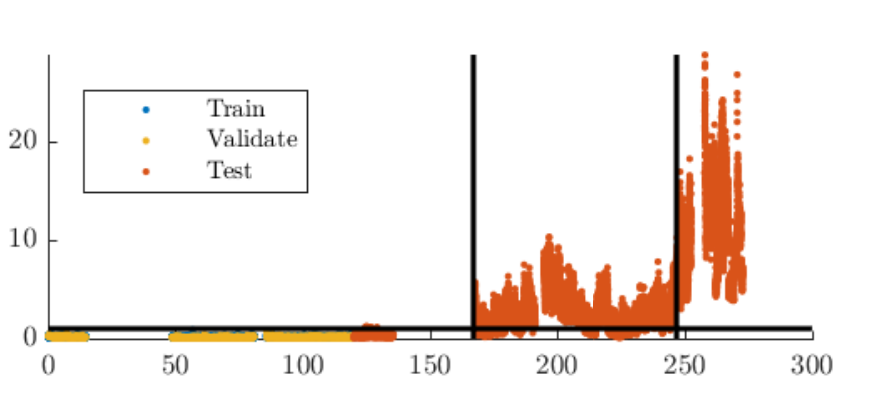}\label{sfig:Dist_1}}	
	\subfloat[ELM (TP: 26.6\% \--- FP: 1.0\%)]
	{\includegraphics[width=8.2cm]{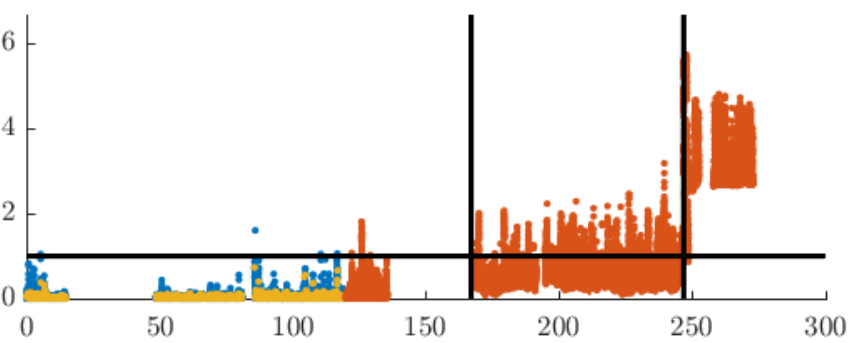}\label{sfig:Dist_2}}\\
	\subfloat[PCA-ELM (TP: 35.6\% \--- FP: 3.2\%)]
	{\includegraphics[width=8.2cm]{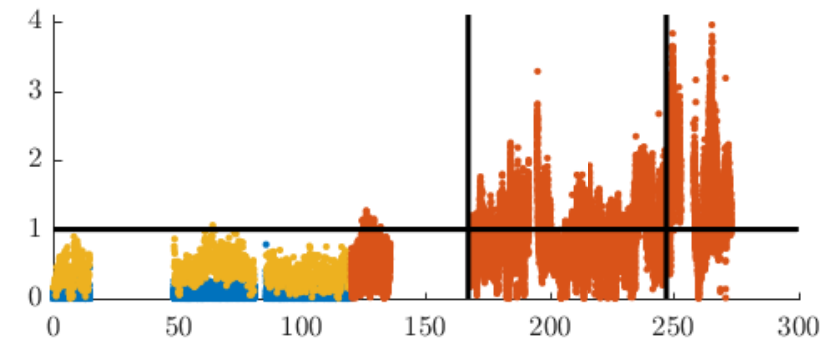}\label{sfig:Dist_3}}	
	\subfloat[SVM (TP: 73.6\% \--- FP: 0.0\%)]
	{\includegraphics[width=8.2cm]{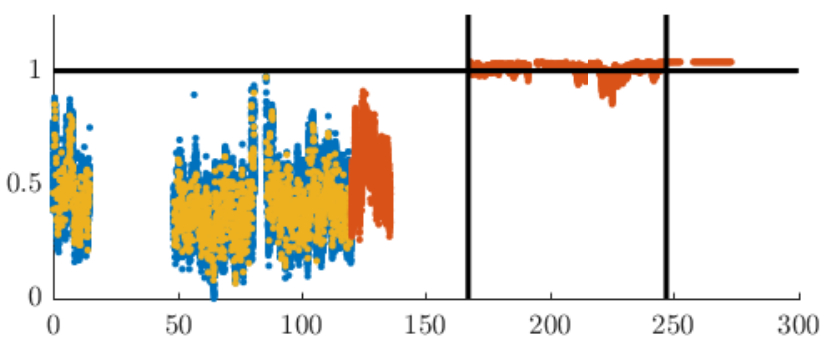}\label{sfig:Dist_4}}\\
	\subfloat[PCA-SVM (TP: 39.8\% \--- FP: 1.2\%)]
	{\includegraphics[width=8.2cm]{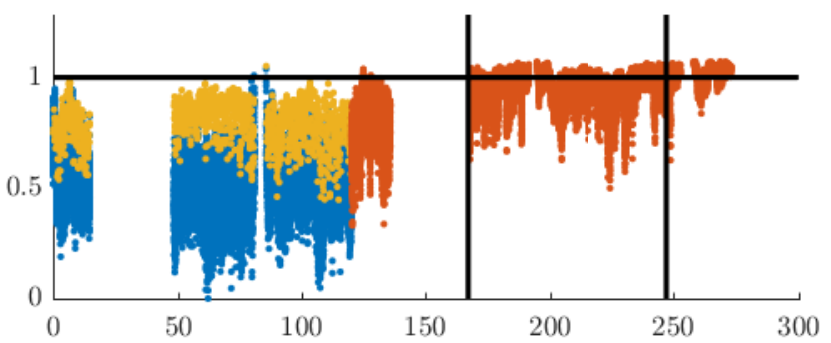}\label{sfig:Dist_5}}	
	\subfloat[DBN (TP: 56.3\% \--- FP: 0.0\%)]
	{\includegraphics[width=8.2cm]{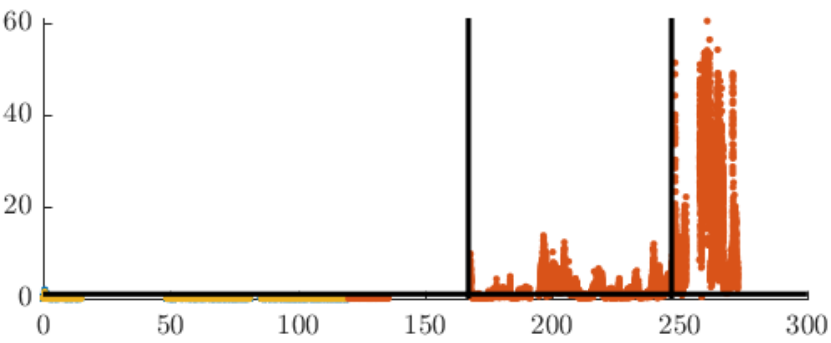}\label{sfig:Dist_6}}
	\caption{Generator Data Case Study:
	\textit{Distance} to normal conditions for the 6 approaches against time (days).
	Blue points represent the training, yellow points the points used for computing the threshold and red points the testing.
	The black horizontal lines correspond to the threshold in Eq.~\eqref{eq:thrd}.
	The two vertical black lines represent $t=169$ days and $t=247$ days.
	The Y-axis is rescaled such that the threshold is 1.}
	\label{fig:Generator_Results}
\end{figure*}

Table~\ref{tbl:rslt_detection_real} synthesises the accuracy, percentage of TP and FP, the magnification coefficient and the training time for the best performing model of each family.
In addition, the distance to the normal class ($\vert 1-Y\vert$) of these models is illustrated in Figure~\ref{fig:Generator_Results}.
In accordance with the results obtained so far, HELM provides the best performance. Its accuracy is higher and it is a robust model, with a strong magnification (Mag=20) and a short training time. The one-class classifier alone performs sensibly worse: numerous FP are raised, TP are scarce, which is consistent with its much lower magnification.
SVM provides good performances with no FP and a good rate of TP, yet its accuracy is smaller than that of HELM and its magnification is very small (Mag$=1.04$). SVM is the least robust model with respect to the threshold definition and is harder to train (300 times slower).
Performing a PCA before the classification also worsened the results. This could be the consequence of the inherent non-linear relationships between the measurements.
The Deep Belief Network has also a lower accuracy compared to HELM. However, it has a stronger magnification.

Based on Fig.~\ref{fig:Generator_Results}, for all models but SVM, the distance to normal class increases after the lower level fault (from day 169 to day 247) and again after the upper level fault (after day 247). The output of the one-class classifier behaves similarly to a health indicator.

	\subsection{HELM versus Expertise}

	Experts who analysed the power plant operation could identify the fault starting at day 169 thanks to eight particular signals. Using these signals as inputs to the ELM and SVM models mimics the traditional machine learning approach using engineered features as input. Doing so, ELM achieves an accuracy of 88\% (TP: 75.1\%, FP: 0.0\%) and a magnification of 20 and SVM an accuracy of 80\% (TP: 60.2\%, FP:0.0\%) and a magnification of 6. This approach brings ELM results very close to those achieved by HELM in the previous section. For SVM, accuracy decreases but the higher magnification allows now to distinguish between lower and higher level faults.

This approach, however, is based on the measurements selected by the experts \textit{a posteriori} to the fault. It could probably detect similar faults in the future but it may miss other types of faults, impacting other measurements. This flaw is not shared by HELM which does not require any \textit{a posteriori} knowledge.

\section{Discussion and Conclusions}
\label{sec:discussion}

Both case studies analysed in this research demonstrate a better performance of HELM compared to that of other methods.
Tested on high dimensional signals with varied characteristics and faults, HELM raises consistent and robust alarms in faulty system conditions. It also enables tracking in time of the evolution of the unhealthy system states. The performance of other tested methods is more impacted by the dimensionality of the input data, the characteristics of the signals and of the faults. The dimensionality reduction step with PCA, for example, has proven in other experiments to be particularly suitable for signals with linear relationships. However in the presence of non-linear relationships, the ability of PCA to extract non-correlated components decreases significantly.
Other deep learning approaches, such as DBN, can also provide excellent results but are usually highly dependent on the selection of hyperparameters. Experimentally, training an efficient DBN necessitates a lot of fine tuning and has proven to be a difficult task.

On both case studies, HELM demonstrates a performance that is either superior or similar to that of the other approaches on all the datasets. It is, in addition, much faster and more robust. The generator case study demonstrated, in addition, that HELM is able to learn the relevant features also from real condition-monitoring signals and to detect the fault at its earliest stage, with a higher accuracy compared to the manually selected features. This is a promising advantage as the dimensionality of the condition monitoring data tends to increase rapidly since more parameters tend to be monitored due to the decreasing costs of sensors. Features are, therefore, harder to engineer manually and feature selection becomes more challenging with the increasing number of features.

HELM integrates a feature learning step which extracts the high-level representations of the condition-monitoring signals and improves the detection accuracy. HELM is able to handle raw condition-monitoring signals and does not require knowledge on the faults as it is trained on normal system state data solely. Its output can be interpreted as a distance to the normal system condition, or a health indicator, and provides, as such, more information over time than the typically applied binary decision process (in cases for which the problem is formulated as a binary classification task). Detecting very early anomalies (months before the faults occur, as in the generator case study) enables to realize savings in terms of operational costs that can reach millions of euros. HELM satisfies, therefore, a lot of the requirements of a good decision support tool for diagnostics engineers.

The proposed approach has shown to be particularly beneficial for industrial applications with high-dimensional signals having complex relationships, experiencing very few but critical faults and lacking representative datasets covering all possible operating conditions and fault types.

For future work, the question of \textit{a priori} hyperparameter setting remains open. While in this study, results were presented for the models that would maximise the accuracies \textit{a posteriori}, it is crucial for diagnostics engineer to know how to set the model prior to the occurrence of faulty system conditions. Nonetheless, we have seen in the simulated case that HELM trained with a good set of parameters is able to detect many different kinds of faults. Therefore, inducing artificial faults on real condition monitoring data could help finding a good set of hyperparameters that will likely be valid for other types of faults.
The investigation of the number of autoencoder layers on the feature learning performance is also left for future works. In this contribution, a single feature learning layer was sufficient to demonstrate a superior performance. Additional feature learning layers could improve results or be necessary if the dimensionality of the dataset is higher than in the present case.

\section*{Acknowledgment}
Work supported by the Swiss Commission for Technology and Innovation under grant number 17833.2 PFES-ES.

\printbibliography

\end{document}